\documentclass[letterpaper, 10 pt, journal, twoside]{ieeetran}
\usepackage{amsmath,amsfonts}
\usepackage{algorithmic}
\usepackage{array}
\usepackage{textcomp}
\usepackage{stfloats}
\usepackage{subcaption}
\usepackage{url}
\usepackage{verbatim}
\usepackage{graphicx}
\usepackage{cite}
\usepackage[utf8]{inputenc} 
\usepackage[T1]{fontenc}    
\usepackage{url}            
\usepackage{booktabs}       
\usepackage{amsfonts}       
\usepackage{nicefrac}       
\usepackage{microtype}      
\usepackage{graphicx} \graphicspath{{figures/}}
\usepackage{amsmath,amssymb,mathbbol}
\usepackage{tabularx,colortbl,multirow,array,makecell}
\usepackage{algorithmic}
\usepackage[linesnumbered,ruled,vlined]{algorithm2e}
\usepackage{acronym}
\usepackage{enumitem}
\usepackage[dvipsnames]{xcolor}
\usepackage[pagebackref,breaklinks,colorlinks,citecolor=gray]{hyperref}
\usepackage{xspace}
\usepackage[skip=3pt,font=small]{subcaption}
\usepackage[skip=3pt,font=small]{caption}
\usepackage[capitalise]{cleveref}
\usepackage[misc]{ifsym}
\usepackage{pifont}

\makeatletter
\DeclareRobustCommand\onedot{\futurelet\@let@token\@onedot}
\def\@onedot{\ifx\@let@token.\else.\null\fi\xspace}

\makeatother

\crefname{algorithm}{Alg.}{Algs.}
\Crefname{algocf}{Algorithm}{Algorithms}
\crefname{section}{Sec.}{Secs.}
\Crefname{section}{Section}{Sections}
\crefname{table}{Tab.}{Tabs.}
\Crefname{table}{Table}{Tables}
\crefname{figure}{Fig.}{Fig.}
\Crefname{figure}{Figure}{Figure}

\makeatletter
\renewcommand{\paragraph}{%
  \@startsection{paragraph}{4}%
  {\z@}{0ex \@plus 0ex \@minus 0ex}{-1em}%
  {\hskip\parindent\normalfont\normalsize\bfseries}%
}
\makeatother

\definecolor{wcolor}{RGB}{251,76,31}

\acrodef{esrp}[ESRP]{Embodied Scene Rearrangement Planning}
\acrodef{srp}[SRP]{Scene Rearrangement Planning}
\acrodef{pomdp}[POMDP]{Partially Observable Markov Decision Process}
\acrodef{vln}[VLN]{Visual Language Navigation}
\acrodef{esrp-pd}[ESRP-PD]{Embodied Scene Rearrangement Planning Paired Dataset}
\acrodef{tamp}[TAMP]{Task and Motion Planning}
\acrodef{sr}[SR]{Success Rate}
\acrodef{osr}[OSR]{Object Success Rate}
\acrodef{rdr}[RDR]{Remaining Distance Ratio}
\acrodef{il}[IL]{Imitation Learning}
\acrodef{rl}[RL]{Reinforcement Learning}
\acrodef{vlm}[VLM]{Vision-Language Model}
\acrodef{ppo}[PPO]{Proximal Policy Optimization}
\acrodef{bc}[BC]{Behavior Cloning}
\acrodef{iou}[IoU]{Intersection over Union}
\acrodef{mlp}[MLP]{Multi-Layer Perceptron}

\newcommand{\benchmark}{ESRP-Bench\xspace}
\newcommand{\rmnum}[1]{\romannumeral #1}

\usepackage[most]{tcolorbox}
\usepackage{xcolor}
\definecolor{mygray}{RGB}{230,230,230}
\begin{document}
\title{Embodied Scene Rearrangement Planning}

\author{Canzhi~Chen$^{*}$, Zan~Wang$^{*}$, Siqi~Zhu$^{*}$, Qi~Wu$^{*}$, Yixuan~Li$^{*}$, and Wei~Liang$^{\dagger}$%
\thanks{Manuscript received: February 14, 2026; Revised: June 20, 2026; Accepted: July 24, 2026. This paper was recommended for publication by Editor Aniket Bera upon evaluation of the Associate Editor and Reviewers' comments. This work was supported by the National Natural Science Foundation of China (NSFC) under Grant No.~62172043.}%
\thanks{$^{*}$C. Chen, Z. Wang, S. Zhu, Q. Wu, and Y. Li contributed equally.}%
\thanks{$^{\dagger}$ W. Liang is the corresponding author.}%
\thanks{All authors are with the School of Computer Science and Technology, Beijing Institute of Technology, Beijing 100081, China (e-mail: chencanzhi@bit.edu.cn; wangzan@bit.edu.cn; zhusiiqii@gmail.com; wuqi@bit.edu.cn; liyixxuan@gmail.com; liangwei@bit.edu.cn).}%
\thanks{Digital Object Identifier (DOI): 10.1109/LRA.2026.3728329.}%
\thanks{\textcopyright~2026 IEEE. Personal use of this material is permitted. Permission from IEEE must be obtained for all other uses, in any current or future media, including reprinting/republishing this material for advertising or promotional purposes, creating new collective works, for resale or redistribution to servers or lists, or reuse of any copyrighted component of this work in other works.}%
}

\markboth{IEEE Robotics and Automation Letters. Preprint Version. Accepted July, 2026}%
{Chen \MakeLowercase{\textit{et al.}}: Embodied Scene Rearrangement Planning}
\maketitle

\begin{abstract}
This paper introduces \acf{esrp}, a novel task requiring embodied agents to rearrange furniture in $3$D scenes to match a target configuration using only egocentric observations and a top-down target layout. Unlike prior rearrangement tasks, \ac{esrp} precludes global state access and introduces mutual object occlusions, reflecting the practical constraints of real-world robotic deployment. These factors make aligning partial egocentric observations with the global target layout particularly challenging for long-horizon planning.
To facilitate research, we present \benchmark, a comprehensive benchmark built on OmniGibson featuring over $5,400$ scene pairs and $8,200$ objects. We define three multi-level metrics to evaluate rearrangement quality and provide four baselines: a hierarchical task-and-motion planning method, a vision-language-model-based method, and two learning-based approaches~(IL and RL).
Experimental results demonstrate that current methods struggle to complete the task efficiently, highlighting \ac{esrp} as a challenging frontier for embodied agents in scene understanding and long-horizon task planning. This work serves as a stepping stone toward deploying intelligent agents in real-world scenarios. Project page: \url{https://pie-lab.cn/ESRP/}.
\end{abstract}


\section{Introduction}
\IEEEPARstart{E}{xisting} studies on rearrangement planning generally fall into two distinct categories.
(\rmnum{1}) The first category focuses on \textit{tabletop rearrangement}, where robotic manipulators are used to reposition small objects within confined workspaces~\cite{king2016rearrangement, labbe2020monte, goyal2022ifor, ren2024neural}. These methods are limited to tabletop-scale and overlook larger scene-level rearrangement, which introduces additional challenges such as long-horizon navigation and spatial reasoning within complex environments.
(\rmnum{2}) Another line of research targets \textit{scene-level object rearrangement}, but typically operates on a simplified 2D plane with access to global observations~\cite{xiong2020motion, wang2024mastering, batra2020rearrangement, wang2020scene}.
However, abstracting the environment to $2$D significantly limits their applicability to embodied agents in realistic $3$D settings, where spatial geometric constraints and partial observations critically affect the feasibility of rearrangement actions.
These limitations highlight a substantial gap between current research efforts and the practical demands of furniture rearrangement in real-world $3$D scenes, such as domestic and hospitality environments.

\begin{figure}[t!]
    \centering
    \includegraphics[width=\linewidth]{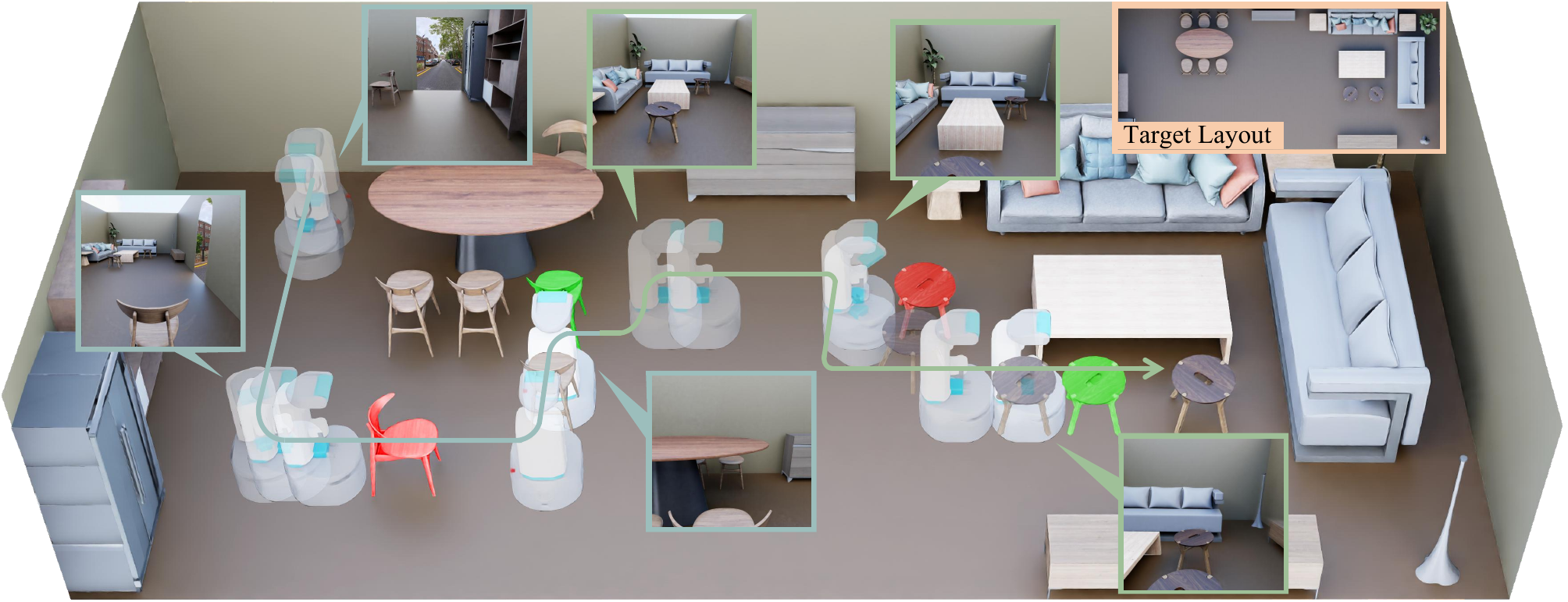}
    \caption{\textbf{Overview of the \ac{esrp} task.} Given a top-down target layout, an embodied agent needs to rearrange furniture in a $3$D scene from an initial configuration~(red) to a target state~(green). The agent operates under realistic constraints, relying only on egocentric observations without access to global state information or ground-truth localization.}
    \label{fig:teaser}
\end{figure}

To bridge these gaps, we introduce \acf{esrp}, a novel task that requires embodied agents to rearrange a $3$D scene from an initial layout to a target configuration, given only egocentric observations and a top-down target layout image, as shown in~\cref{fig:teaser}. Compared to prior works, \ac{esrp} introduces three distinct challenges:
\begin{itemize}[leftmargin=*,nolistsep,noitemsep]
    \item \textbf{Partial Observability:} Relying on egocentric observations significantly increases task difficulty. The policy must align partial, first-person views with the target layout and infer spatial information solely from local perception.
    \item \textbf{Complex Inter-Object Dependencies:} Manipulating large furniture items transforms the task into a long-horizon planning problem. The agent must generate plausible action sequences while reasoning about inter-object dependencies to avoid physical deadlocks and mutual occlusions.
    \item \textbf{Dynamic Scene Evolution:} The rearrangement process dynamically transforms the $3$D scene, imposing heightened demands on the agent's perception, localization, and navigation capabilities within a non-stationary environment.
\end{itemize}

Correspondingly, we introduce \benchmark, a comprehensive benchmark built upon OmniGibson~\cite{li2023behavior} and providing the systematic evaluation of \ac{esrp}. \benchmark comprises over $5,400$ fully furnished indoor scenes featuring more than $8,200$ movable furniture items sourced from the $3$D-FRONT dataset~\cite{fu20213d}, covering diverse room types and furniture categories. Each scenario consists of three key components: (\rmnum{1}) an initial scene layout configuration, (\rmnum{2}) a target layout represented as a top-down view image, and (\rmnum{3}) expert demonstration trajectories to support imitation learning. Furthermore, \benchmark introduces three multi-level metrics for rigorous assessment and facilitates evaluation across varying difficulty levels, determined by the number of objects to be rearranged.

We develop four distinct baseline policies to tackle \ac{esrp}, categorized into three paradigms.
(\rmnum{1}) \textbf{Learning-based approaches:} We implement both an \ac{il} and an \ac{rl} agent, which leverage egocentric observations and a top-down view of the goal layout. The \ac{rl} baseline employs \ac{ppo}~\cite{schulman2017proximal}, while the \ac{il} baseline utilizes \ac{bc}~\cite{gavenski2024survey}. The \ac{ppo} method adopts a vision encoder followed by a single-layer LSTM to process egocentric observations and maintain temporal memory, whereas \ac{bc} adopts a Diffusion Policy~\cite{chi2023diffusion} that predicts future action sequences through an iterative denoising process conditioned on visual observations and the gripper state.
(\rmnum{2}) \textbf{Foundation-model-based approach:} We additionally introduce \textit{ESRP-VLM}, a multimodal ReAct~\cite{yao2023react} agent built on top of a pretrained \ac{vlm}, which examines whether off-the-shelf generalist priors transfer to \ac{esrp} via in-context prompting alone.
(\rmnum{3}) \textbf{Planning-based approach:} We additionally deploy a baseline that exploits privileged global information. The \textit{ESRP-PLAN} baseline implements a \ac{tamp} framework that hierarchically decomposes the problem: a task planner determines the optimal rearrangement order, while a motion planner computes collision-free paths for fetching and placing each object.

Despite employing diverse baseline models, experimental results indicate that performance remains suboptimal. For instance, even the strongest baseline achieves a success rate of only $30.20\%$, plummeting to nearly $0\%$ in complex scenarios, underscoring the significant challenges inherent to \ac{esrp}.

We summarize our contributions as threefold: (\rmnum{1}) We propose \ac{esrp}, a novel and practical task that requires embodied agents to rearrange furniture within $3$D scenes using only egocentric observations. (\rmnum{2}) We introduce a comprehensive benchmark, \benchmark, comprising over $5,400$ initial-target scene layout pairs and three multi-level evaluation metrics that provide a rigorous assessment of rearrangement quality. (\rmnum{3}) We provide four baseline models as a first attempt to tackle \ac{esrp}, accompanied by comprehensive experimental results. These experiments reveal that different paradigms face distinct drawbacks in scene understanding and long-horizon task planning, highlighting key challenges and promising future research directions.

\section{Related Work}
\noindent\textbf{Rearrangement Planning}
Rearrangement entails reconfiguring objects from an initial state to a target layout~\cite{weihs2021visual}. A primary stream of research focuses on manipulating small objects, employing either planning-based methods~\cite{king2016rearrangement, labbe2020monte} or learning-based policies~\cite{ren2024neural, huang2024efficient}. However, these approaches are typically restricted to constrained workspaces and do not scale well to room-level tasks. Conversely, research addressing larger items, such as furniture~\cite{wang2020scene, wei2023lego, wang2024mastering}, often relies on simplified $2$D planar abstractions. This simplification overlooks intricate $3$D spatial relationships and vertical geometric constraints, limiting the transferability of these methods to real-world deployment. To bridge this gap, we propose a task that requires rearranging furniture within fully $3$D scenes using only egocentric perception, marking a significant step toward practical embodied AI applications.

\noindent\textbf{Task and Motion Planning}
\ac{tamp} integrates high-level task planning with low-level motion planning, enabling robots to generate executable actions for complex, long-horizon tasks. Existing \ac{tamp} research has developed diverse approaches, including optimization-based, sampling-based~\cite{orthey2023sampling}, and learning-based methods~\cite{li2025flona}. However, \ac{esrp} presents unique challenges that exceed the capabilities of standard \ac{tamp} formulations. Specifically, it introduces two critical difficulties that strain existing planning methodologies:
(\rmnum{1}) Long-Horizon Task Complexity: Furniture rearrangement necessitates a sequence of interdependent actions over long horizons, surpassing the complexity of simpler manipulation tasks.
(\rmnum{2}) $3$D Spatial Reasoning Requirements: \ac{esrp} demands intricate $3$D spatial understanding within the planning process, requiring precise object positioning, robust collision avoidance, and comprehensive reasoning about scene geometry.
In this work, we systematically analyze the challenges posed by \ac{esrp} through the evaluation of these four baseline approaches.


\section{\benchmark}

\begin{figure}[t!]
    \centering
    \includegraphics[width=\linewidth]{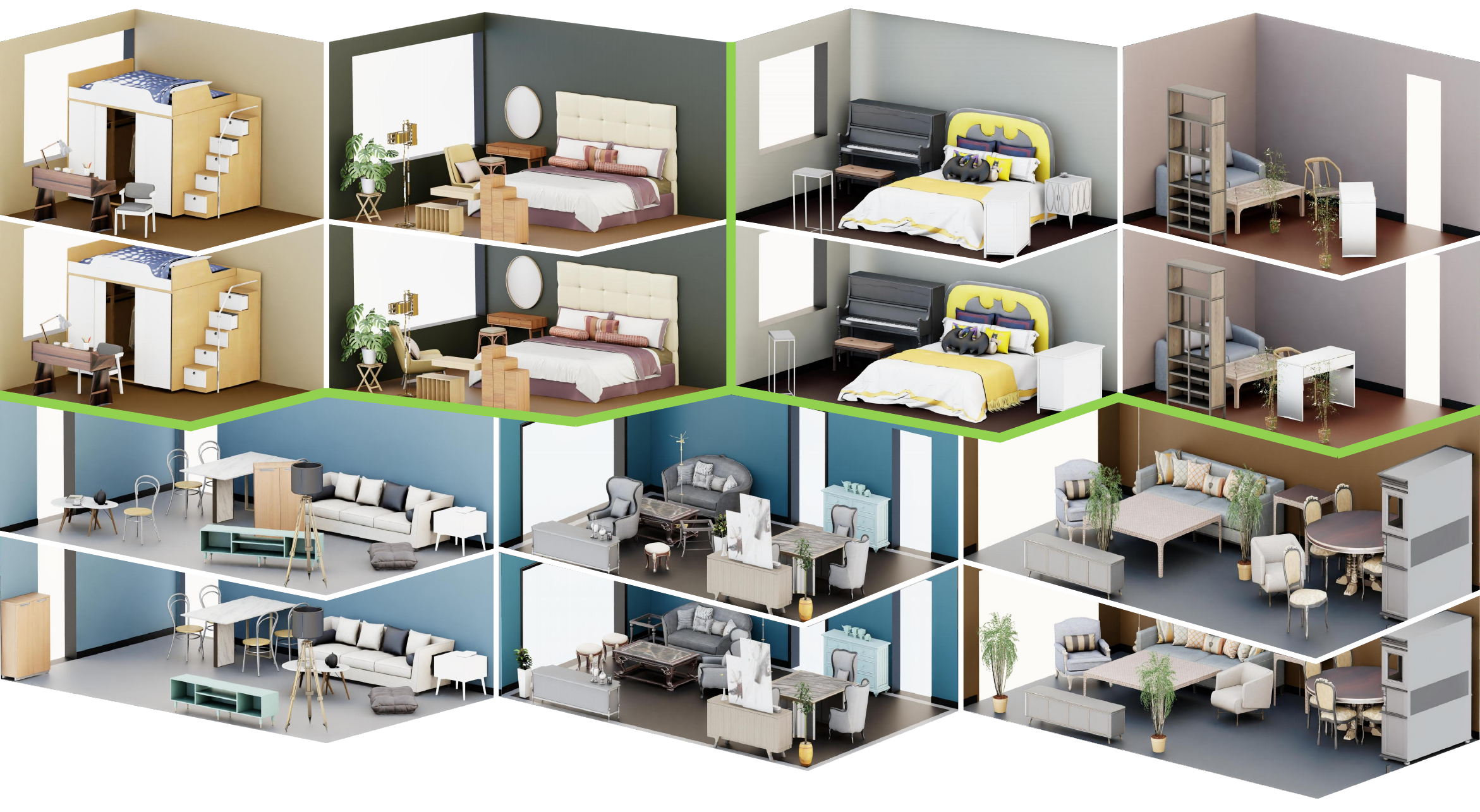}
    \caption{\textbf{Initial-target scene layout pairs.} 
    We group scene layouts into three difficulty levels (as divided by green lines), with the first row showing the initial layout and the second row displaying the corresponding target layout. The top left shows \textbf{easy-level} scenes, where only $1$ furniture object needs to be rearranged; the top right presents \textbf{medium-level} scenes with $2 \sim 3$ objects to be rearranged; and the bottom shows \textbf{hard-level} scenes with $4 \sim 6$ objects to be rearranged. The dataset covers a wide range of room types, including bedrooms, study rooms, and living rooms, along with diverse furniture types, such as chairs, tables, and plants.}
    \label{fig:benchmark}
\end{figure}

\subsection{Task Definition}
\label{sec: task}
\ac{esrp} defines a task in which an agent needs to reconfigure an unseen 3D indoor scene to match a target layout specified by a single top-down image. Relying solely on first-person views, the agent is required to: (\rmnum{1}) comprehend the current scene state through egocentric observations, (\rmnum{2}) formulate an efficient rearrangement strategy, and (\rmnum{3}) execute a sequence of actions to precisely reposition furniture items. The task is deemed successful only when all rearrangeable objects are correctly placed at their designated goal positions.

\noindent\textbf{Observations} 
At each timestep, the \ac{esrp} agent receives an observation tuple comprising three components:
(\rmnum{1}) an egocentric RGB image $I^{ego}$ ($128\times128$) capturing the agent's current first-person view;
(\rmnum{2}) a top-down RGB image $I^{g}$ ($128\times128$) depicting the target furniture layout;
and (\rmnum{3}) a binary proprioceptive indicator $g \in \{0,1\}$ denoting whether the agent is currently grasping an object.

\noindent\textbf{Actions} 
We define the action space $\mathcal{A}=\,$\{\texttt{move\_forward}, \texttt{move\_backward}, \texttt{turn\_left}, \texttt{turn\_right}, \texttt{fetch}, \texttt{release}\}. The first four actions constitute the navigation primitives, while the latter two are object manipulation primitives. To balance planning efficiency with control precision, we implement discrete motion primitives: rotations are fixed in-place turns of $22.5^\circ$, and translations are defined as fixed steps of $0.25$m.

\noindent\textbf{Success Conditions}
We define episode success as the successful placement of all rearrangeable objects into their target configurations. Quantitatively, an object is considered correctly positioned if the \ac{iou} between its current $2$D bounding box and the target bounding box exceeds $0.3$ on the horizontal plane. This threshold provides a robust geometric tolerance to accommodate minor control noise, while strictly ensuring that the final scene layout maintains semantic fidelity to the target.

\noindent\textbf{Termination Conditions} 
An episode terminates upon satisfying either of the following conditions: (\rmnum{1}) the agent exhausts the maximum allowable timestep horizon $L$, or (\rmnum{2}) all rearrangeable objects are successfully positioned at their target locations. The former constitutes a timeout failure, while the latter signifies successful task completion.

\begin{figure*}[t!]
    \centering
    \begin{subfigure}{0.5\linewidth}
        \centering
        \includegraphics[width=\linewidth]{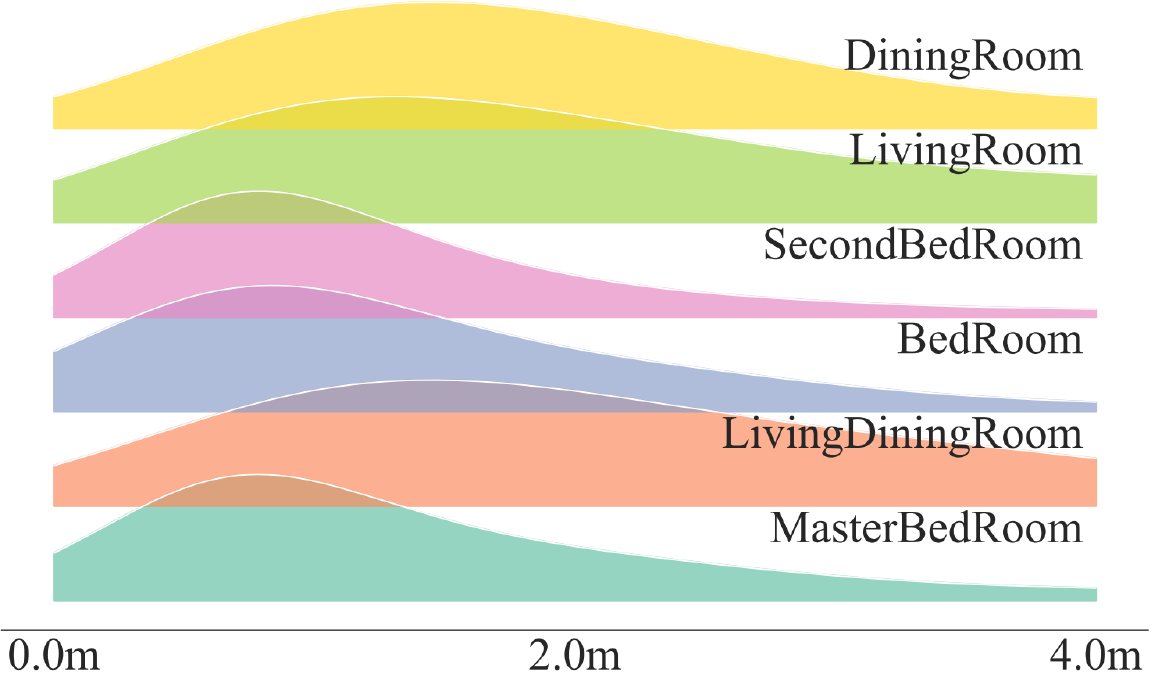}
        \caption{Distribution of Scene Clutter Level}
        \label{fig:object distribution}
    \end{subfigure}\hfill%
    \begin{subfigure}{0.5\linewidth}
        \centering
        \includegraphics[width=\linewidth]{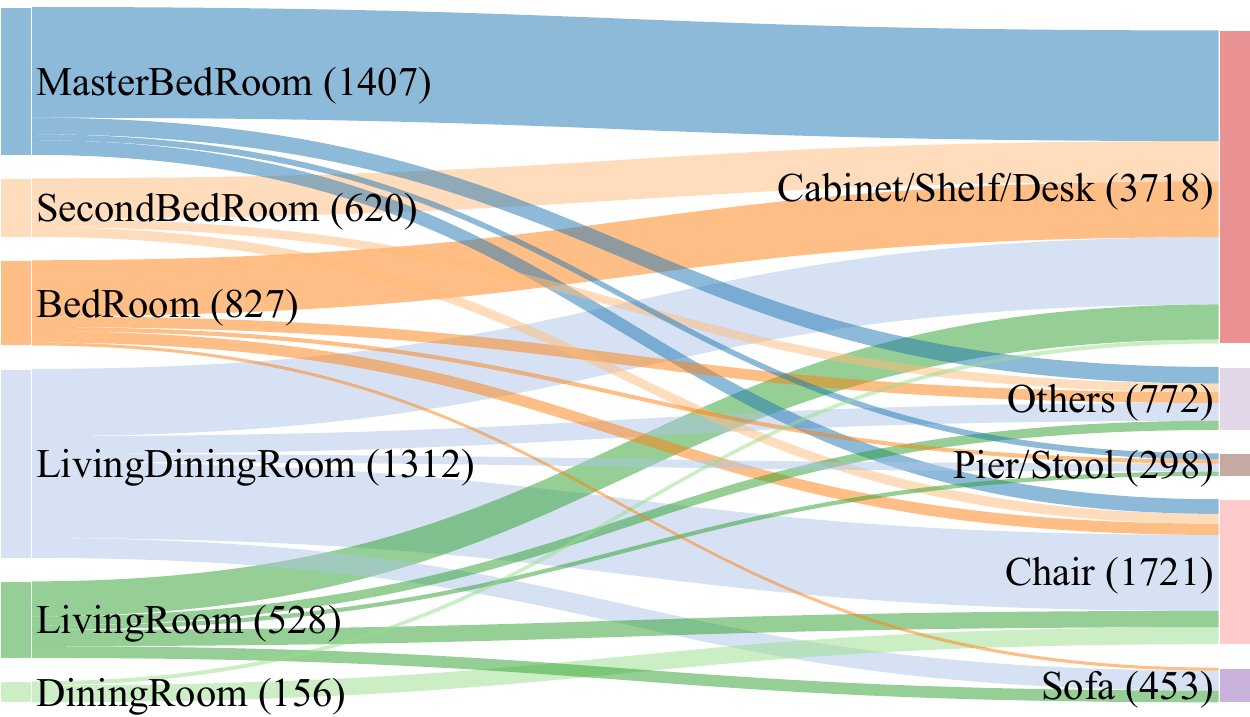}
        \caption{Distribution of Rooms and Objects}
        \label{fig:room distribution}
    \end{subfigure}
    \caption{\textbf{Dataset distribution.} (a) The scene clutter level in the dataset is defined as the sum of L$2$ distances between the initial and target positions of all rearrangeable objects. Formally, more challenging scenes have higher clutter levels. (b) Distribution of rearrangeable objects across different room types in the dataset. We report only the top six room categories in the dataset, excluding those that constitute only a small fraction. Notably, these top six categories account for over 80\% of all rooms.}
    \label{fig:dataset}
\end{figure*}

\subsection{Dataset}
\label{sec: dataset}
We introduce a comprehensive benchmark dataset tailored for \ac{esrp}, derived from the $3$D-FRONT dataset~\cite{fu20213d}. This benchmark provides paired configurations with annotated initial and target layouts. In total, the dataset comprises $5,495$ distinct scenes spanning $24$ room types, with each scene containing between $1$ and $6$ rearrangeable objects from common furniture categories.
\cref{fig:benchmark} illustrates examples of these paired scenes across three difficulty levels: \textit{Easy} ($1$ object), \textit{Medium} ($2\text{--}3$ objects), and \textit{Hard} ($4\text{--}6$ objects). This stratification ensures the dataset encompasses a broad range of indoor layouts, facilitating evaluation across varying degrees of spatial complexity and object interaction.
Additionally, \cref{fig:dataset} presents two complementary views of the dataset statistics: the distribution of scene clutter levels and the frequency of rearrangeable objects across different room types.

\noindent\textbf{Scene and Object Curation}
To ensure task feasibility, we applied several filtering criteria during dataset construction. 
First, we excluded rooms with insufficient navigational space for a furniture-carrying agent, preventing scenarios where physical constraints would make rearrangement impossible. 
Second, we selected furniture items with dimensions that allow effective agent manipulation, specifically choosing pieces slightly smaller than the agent to maintain maneuverability. 
Finally, to balance task complexity with practical solvability, we constrained each scene to contain at most $6$ rearrangeable objects. 
These design choices create a challenging yet tractable benchmark that focuses on core rearrangement skills rather than navigation in extremely confined spaces.

\noindent\textbf{Target Layout Acquisition}
We leverage the professionally designed layouts from the 3D-FRONT dataset~\cite{fu20213d} as the target layouts. We then capture $128\times128$ RGB images from a top-down view of these scenes to serve as the goal images. To represent the target layouts, we render $128 \times 128$ top-down orthographic RGB images, offering a complete spatial reference for agents. These images serve as a concise and comprehensive depiction of the desired furniture configuration, aiding agents in planning and executing their rearrangement tasks effectively.

\noindent\textbf{Initial Layout Generation}
To ensure the rearrangement task is feasible, we generate the initial layout by perturbing the target layout through a rule-based simulation of object movements. Starting from the goal configuration, the robot iteratively moves each rearrangeable object, taking random steps forward/backward and turning, until it reaches a non-overlapping, collision-free position. This backward construction ensures, to the greatest extent possible, that a feasible solution exists between the initial and target layouts. Similar initialization strategies have been adopted in prior work~\cite{wei2023lego, wang2020scene}, but they typically ignore the robot’s embodiment, leading to object placements that are infeasible due to collisions during execution. 

\noindent\textbf{Expert Trajectories}
To facilitate the development of \acl{il} approaches for \ac{esrp}, we use the reverse trajectories in the initial layout generation process as expert demonstrations. These trajectories capture rearrangement sequences executed by our rule-based agent, offering researchers high-quality behavioral data for supervised learning, behavioral cloning, and other \ac{il} paradigms.

\noindent\textbf{Statistics of Dataset}
We present comprehensive statistics of our proposed dataset. We analyze the distribution of room types and object categories, distinguishing between fixed furniture elements and rearrangeable objects to provide a detailed characterization of the dataset composition.

\begin{figure}[t!]
    \centering
    \includegraphics[width=0.8\linewidth]{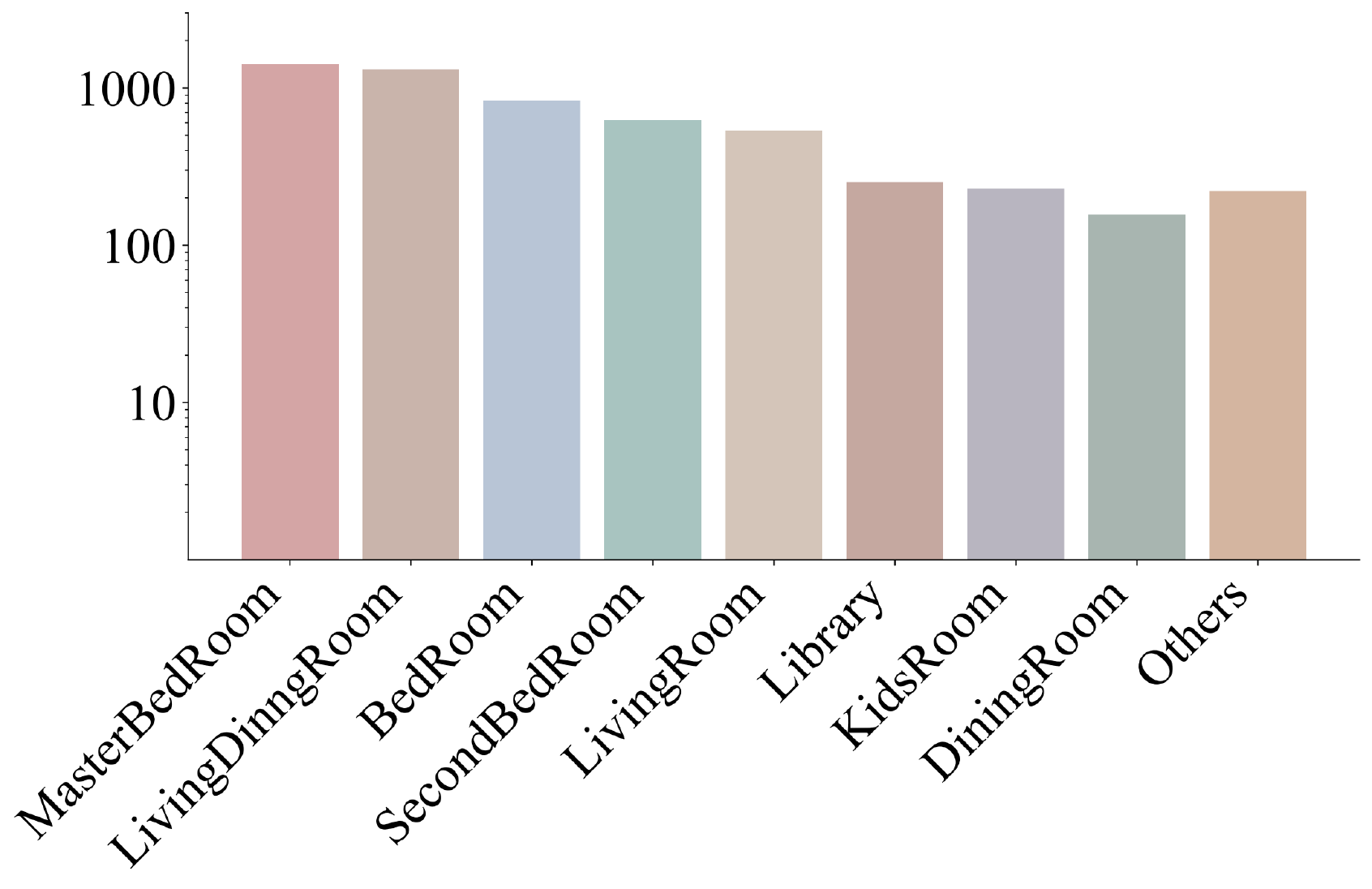}
    \caption{\textbf{Distribution of room types.} The distribution of room categories is visualized on a logarithmic scale to account for class imbalance. Our dataset encompasses a diverse array of room types, comprehensively covering common household environments.}
    \label{fig:datasetstat}
\end{figure}

\cref{fig:datasetstat} illustrates the distribution of room types within the dataset, which encompasses $24$ distinct categories, with $8$ types represented by more than $100$ instances each. Our dataset predominantly targets residential environments where furniture organization is most relevant; notably, \textit{MasterBedroom} and \textit{LivingDiningRoom} constitute the largest proportions.
We further analyze the composition of movable furniture, identifying a total of $8,213$ movable objects. These are categorized into: Cabinet/Shelf/Desk ($3,718$), Chair ($1,721$), Sofa ($453$), Pier/Stool ($298$), and Others ($772$). 


\subsection{Metrics}
\label{sec: metrics}
To quantitatively assess performance on the \ac{esrp} benchmark, we introduce a set of comprehensive evaluation metrics. Each method is evaluated across $N$ independent episodes, with results aggregated by computing the mean performance for each metric. This approach ensures robust statistical assessment while capturing different aspects of rearrangement efficacy.

\begin{itemize}[leftmargin=*,nolistsep,noitemsep]
\item \textbf{\ac{sr}:} \ac{sr} reflects the entire episode success rate and is computed as $\text{SR} = \frac{1}{N}\sum_{i=1}^{N}S_i$, where $S_i$ is a binary indicator of success in episode $i$. An episode is considered successful if every object reaches its goal.
\item \textbf{\ac{osr}:} To quantify the effectiveness of the rearrangement process, we introduce the \ac{osr}, which measures the proportion of objects correctly placed in their target positions. The OSR is formulated as:
$\text{OSR} = \frac{1}{N} \sum_{i=1}^{N} {M_i^{\text{succ}}} / {M_i}$, 
where $M_i$ is the number of objects that need to be rearranged in episode $i$ and $M_i^{\text{succ}}$ is the number of objects successfully rearranged by the end of episode $i$.
\item \textbf{\ac{rdr}:} To evaluate the proximity of the final arrangement to the goal configuration, we introduce the \ac{rdr}, defined as:
$\text{RDR} = \frac{1}{N} \sum_{i=1}^{N} ({D_i^{\text{end}}} / {D_i^{\text{start}})}$,
where $D_i^{\text{start}}$ and $D_i^{\text{end}}$ denote the summed Euclidean (L$2$) distances between all objects and their target positions at the start and end of episode $i$, respectively. Lower RDR values indicate better performance, with values approaching zero representing perfect alignment between all furniture items and their target positions in the goal configuration.
\end{itemize}

\begin{figure*}[ht!]
    \centering
    \includegraphics[width=\linewidth]{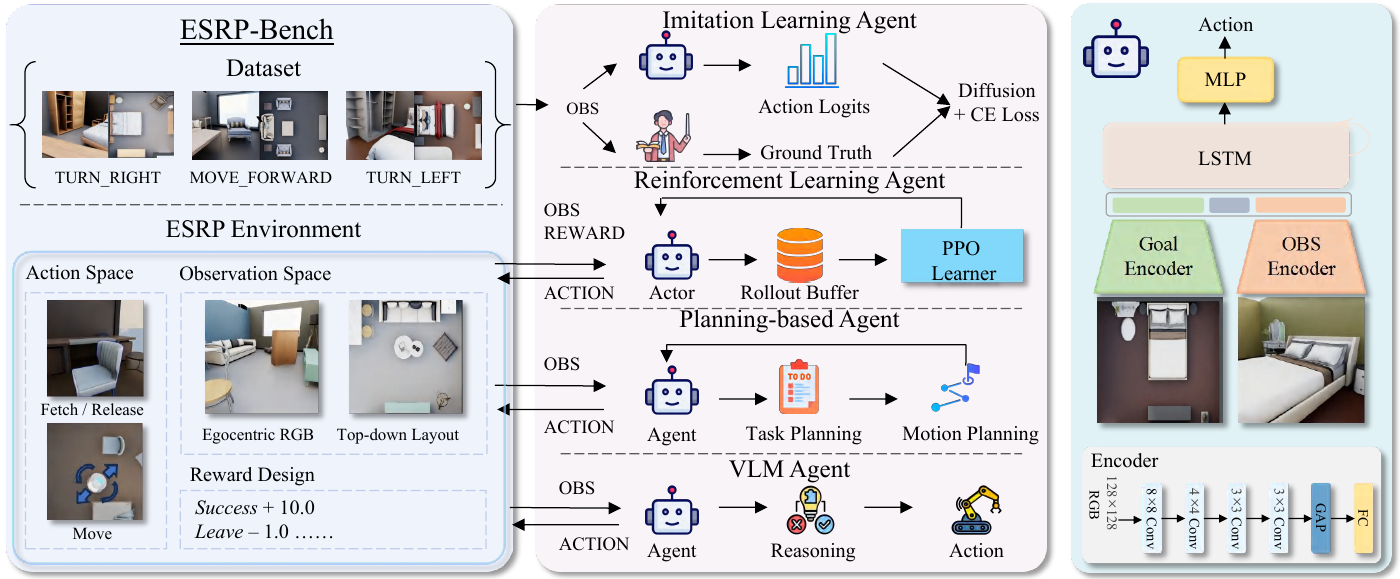}
    \caption{\textbf{Pipeline overview.} (\textbf{Left}) \benchmark\ consists of the dataset and the \ac{esrp} environment. 
    (\textbf{Middle}) Four baseline models are: (1) \ac{il} agent uses expert trajectories from the dataset to train a policy that imitates expert behavior; (2) \ac{rl} agent learns a policy by interacting with the \ac{esrp} environment; (3) foundation-model-based VLM agent prompts a pretrained \ac{vlm} in a ReAct loop without task-specific training; and (4) planning-based agent utilizes global information to perform hierarchical task and motion planning.
    (\textbf{Right}) The model architectures for \ac{rl} are shown, consisting of a CNN for visual perception and an LSTM module for memory. The gray box represents the embedding of the grasping state indicator $g$.}
    \label{fig:pipeline}
\end{figure*}

\subsection{Simulation}
\label{sec: sim}
We implement \ac{esrp} within OmniGibson \cite{li2023behavior}, a physics-based simulation framework built on NVIDIA Isaac Sim.

\noindent\textbf{Robot Setup}
Our benchmark employs the embodiment developed from the Fetch robot, in which the robotic arm is removed, and a camera is mounted overhead.

\noindent\textbf{Action Execution} 
While our primary focus lies in high-level planning and decision-making, we strictly enforce physical fidelity through rigorous collision detection. Adopting the methodology established by HomeRobot~\cite{yenamandra2023homerobot}, we implement continuous collision checking against the full $3$D meshes of the robot, furniture, and environmental structures. This process guarantees trajectory feasibility and ensures the physical plausibility of every executed action.

To streamline simulation, we abstract the low-level grasping mechanics by instantaneously attaching objects to the agent upon interaction. This simplification allows us to prioritize high-level planning over control fidelity, aligning with established methodologies in rearrangement~\cite{weihs2021visual} and task planning~\cite{mirakhor2024task}. Specifically, the agent is capable of fetching objects within a $1.0$\,m radius from its center. This interaction range is physically grounded; considering the robot's diameter of $0.6$\,m and the substantial dimensions of the target furniture, it realistically approximates the reach required for manipulating large objects in real-world scenarios.

\subsection{Baselines}
\label{sec: Methods}

We design four baselines spanning three paradigms~(\cref{fig:pipeline}): an \ac{il} agent \textit{ESRP-BC}; a \ac{rl} agent \textit{ESRP-PPO}; a foundation-model agent \textit{ESRP-VLM} that prompts a pretrained \ac{vlm} in a ReAct~\cite{yao2023react} loop; and a planning-based agent \textit{ESRP-PLAN} that uses hierarchical \ac{tamp} with privileged global state.


\noindent\textbf{\textit{ESRP-BC}}
We deploy a conditional diffusion policy~\cite{chi2023diffusion} that
predicts a sequence of future actions $a_{t:t+h-1}$, where $t$ is the current
environment step and $h$ is the prediction horizon. At each step $t$, the policy is conditioned on the vector $c_t$ by concatenating
features from two ResNet-50~\cite{he2016deep} encoders (initialized from pretrained weights) applied to the egocentric
observation history $I^{ego}_{t-n+1:t}$, and the goal image $I^g_t$, together with embeddings of the
gripper-state history $g_{t-n+1:t}$, and projecting the result with an \ac{mlp}, where $n$ is the history length.

During training, the expert future action sequence is encoded as one-hot vectors
$x^0_{t:t+h-1}$, where the superscript $0$ denotes the clean sample. We sample a
diffusion step $k$ and add Gaussian noise,
$x^k_{t:t+h-1}=\sqrt{\bar{\alpha}_k}x^0_{t:t+h-1}+\sqrt{1-\bar{\alpha}_k}\epsilon, \epsilon \sim \mathcal{N}(0,I),$
where $x^k_{t:t+h-1}$ is the noisy action sequence and
$\bar{\alpha}_k$ is the cumulative noise coefficient. A conditional $1$D U-Net predicts the noise as
$\hat{\epsilon}=\epsilon_\theta(x^k_{t:t+h-1}, k, c_t).$
The objective combines diffusion denoising with a cross-entropy loss on the denoised discrete action logits:
$\mathcal{L}(\theta)=\left\|\epsilon -\epsilon_\theta(x^k_{t:t+h-1}, k, c_t)\right\|_2^2-\lambda_{CE}\sum_{\tau=t}^{t+h-1}\log p_\theta(a_\tau \mid \hat{x}^0_\tau, c_t),$
where $\tau$ indexes future actions and $\hat{x}^0_\tau$ denotes the
denoised logits at step $\tau$.

\noindent\textbf{\textit{ESRP-PPO}} 
Our \ac{rl} policy uses two visual encoders for $I^{ego}_t$ and $I^g_t$, a learnable embedding for the gripper state $g_t$, and a single-layer LSTM with context length $l$ that produces a hidden state $h_t$, which an \ac{mlp} maps to the next action.
Following prior scene-level rearrangement work~\cite{wang2020scene, mirakhor2024task}, we use a dense reward composed of: a $+10$ \textit{Success} reward upon full task completion; a $\pm 1$ \textit{Arrival/Leave} signal when an object is released at, or displaced from, its goal; a \textit{Potential} term $r_{\text{potential}}\cdot(D_{t-1}-D_{t})$ with $D_t$ the L$2$ distance from the held object to its target; a $+0.01$ per-step \textit{Grasping} bonus to discourage premature releases; and a \textit{Living} term (set to $0$). Collisions incur no explicit penalty: any colliding action is rejected at the simulator level, so the agent simply forfeits the corresponding shaping rewards.

\noindent\textbf{\textit{ESRP-VLM}}
This baseline builds a multimodal ReAct~\cite{yao2023react} agent on top of the pretrained Qwen3-VL-2B-Instruct~\cite{qwen3vl} \ac{vlm}. At each step, it receives $I^{ego}_t$ and $I^g_t$ as visual inputs and a textual context that encodes $g_t$ and a brief recent action history; following the ReAct paradigm, it alternates between a short reasoning trace and a discrete action $a_{t+1}\in\mathcal{A}$ via in-context prompting, without task-specific fine-tuning. This baseline represents the rapidly evolving family of foundation-model-based embodied agents.

\noindent\textbf{\textit{ESRP-PLAN}} 
Our planning-based baseline assumes perfect perception (complete occupancy map, exact poses of all movable and unmovable objects, and target locations) and uses a hierarchical \ac{tamp} framework. The task planner determines the rearrangement order, and for each object the motion planner runs a four-phase routine: (\rmnum{1}) BFS to the target object on a state space discretized at $0.05$\,m and $\pi/32$\, rad, (\rmnum{2}) fetch, (\rmnum{3}) A* to the goal with Euclidean heuristic, and (\rmnum{4}) release. Collisions are checked against 2D polygonal obstacle representations, and a goal IoU threshold of $0.3$ is used for verification.

\section{Experiments}

\subsection{Experimental Setup}  
\noindent\textbf{Implementation Details} 
For the \ac{rl} model, we employ \ac{ppo} with a learning rate of $1.5 \times 10^{-4}$ and a value function loss coefficient of $5 \times 10^{-3}$. Training is performed with a batch size of $2048$, using minibatches of $64$ for gradient updates. The history context length $l$ is fixed at $64$ time steps. We train the policy for $6$ million environment steps until convergence. 

For the \ac{il} model, we train for 500 epochs with $50$ diffusion timesteps, using AdamW with a learning rate of $10^{-4}$, weight decay of $10^{-6}$, a batch size of $1024$, a cross-entropy weight of $\lambda_{CE}=0.2$, the prediction horizon $h=8$, and the history length $n=4$. At inference time, we use DDIM sampling with
$D=16$ denoising steps. The policy predicts an action sequence of length $8$, while only the first $6$ actions are executed, with a $0.1$ probability of taking a random action at each step. This configuration is empirically found to yield the best performance.

For the planning-based model, we use BFS for navigation to objects with a maximum of $1000$ iterations and A* for navigating to the goal with the same iteration limit. The state space is discretized with $0.05$m grid spacing for positions and $\pi/32$ radians for orientations. Task planning uses an exhaustive search over object orderings when feasible, selecting the ordering that places the most objects successfully.

The IL model is trained on an NVIDIA A100 GPU, while all other training and inference experiments are conducted on an NVIDIA RTX 4090 GPU.

\noindent\textbf{Dataset}
We split our dataset into training and test sets comprising $90\%$ and $10\%$ of the scenes, respectively. The training split is used for both \ac{rl} and \ac{il} models' training.


\begin{figure*}[t!]
    \centering
    \includegraphics[width=\linewidth]{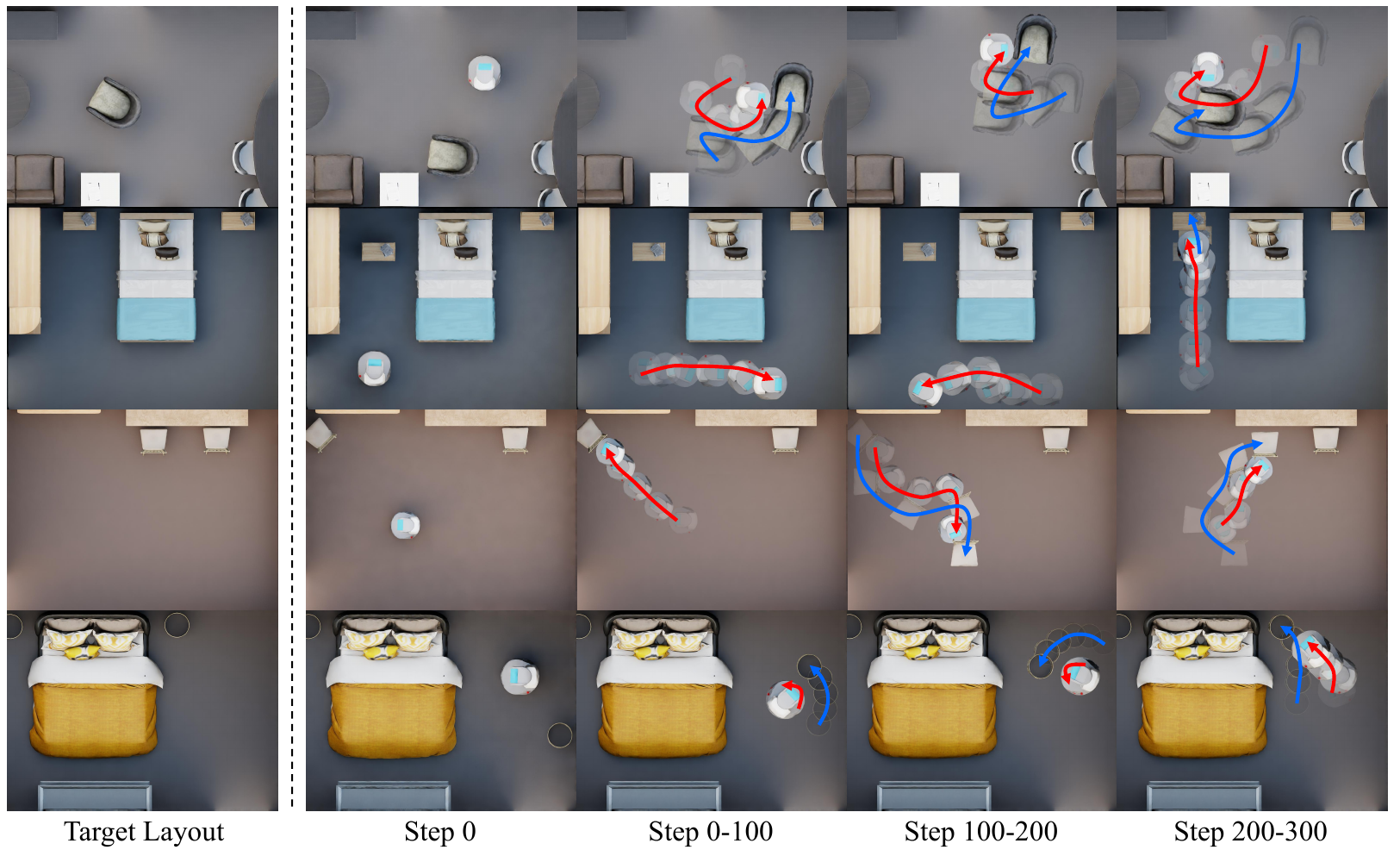}
    \caption{\textbf{Examples of successfully rearranged scenes.} The agent successfully moves the objects to match the target layout. The red curve represents the agent’s trajectory, while the blue curve depicts the object’s trajectory.}
    \label{fig:results}
\end{figure*}

\begin{figure*}[t!]
    \centering
    \includegraphics[width=\linewidth]{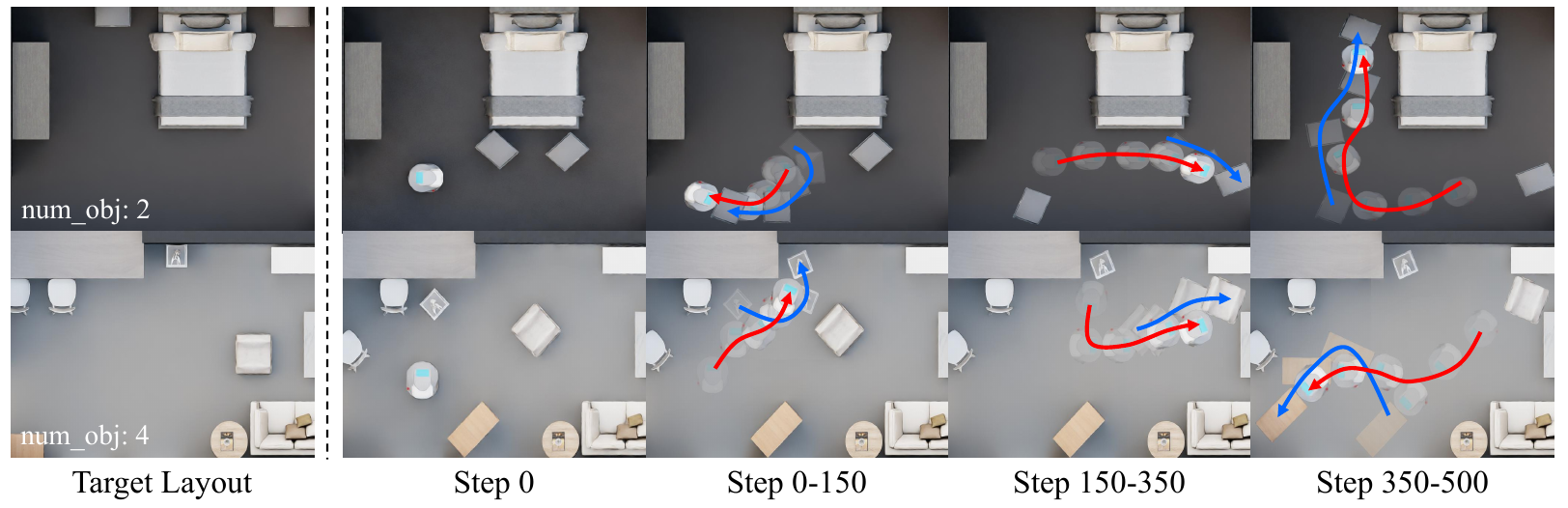}
    \caption{\textbf{Failure cases.} As the number of rearrangeable objects increases, the task becomes more challenging due to its long-horizon complexity. In such cases, the agent rearranges the scene to approximate the target layout.}
    \label{fig:resultf}
\end{figure*}

\subsection{Results}
\label{sec: result}

\noindent\textbf{Overall Rearrangement Results}\label{para: overall_result} 
\Cref{tab:results} demonstrates that: (\rmnum{1}) \textit{ESRP-PLAN} outperforms all other baselines across all three metrics. This advantage arises from its access to the global environment state, encompassing the current and target positions of all rearrangement objects, the agent’s pose, and the full obstacle occupancy of the scene, which enables explicit planning for navigation, object fetching, and obstacle avoidance. In contrast, learning-based approaches rely on partial observations and therefore exhibit weaker performance. (\rmnum{2}) \textit{ESRP-PPO} consistently outperforms \textit{ESRP-BC} across all three metrics. While \textit{ESRP-BC} achieves comparable \ac{sr} and \ac{osr} with \textit{ESRP-PPO}, indicating that behavior cloning has learned a certain degree of rearrangement capability from expert demonstrations, \textit{ESRP-BC} performs worse in \ac{rdr}, suggesting that errors accumulate and are amplified during long-horizon execution. In contrast, \textit{ESRP-PPO} exhibits more comprehensive performance and maintains more stable behavior over long action sequences. (\rmnum{3}) \textit{ESRP-VLM}, despite leveraging a pretrained vision-language foundation model, achieves performance comparable to \textit{ESRP-BC} but does not surpass \textit{ESRP-PPO}, indicating that off-the-shelf foundation-model priors alone are insufficient for the long-horizon spatial reasoning required by \ac{esrp}.

\begin{table}[t!]
    \centering
    \caption{\textbf{Performance of the baseline methods on the test set.}}
    \label{tab:results}
        \begin{tabular}{lccc} 
            \toprule
            Baseline & SR $\uparrow$ & OSR $\uparrow$ & RDR $\downarrow$\\
            \midrule
            ESRP-PPO    & $20.18\%$ & $20.35\%$ & $0.9845$  \\
            ESRP-BC     & $19.09\%$ & $20.31\%$ & $1.0170$  \\
            ESRP-VLM    & $17.64\%$ & $18.97\%$ & $0.9830$  \\
            ESRP-PLAN   & $\textbf{30.20}\%$ & $\textbf{36.59}\%$ & $\textbf{0.6916}$ \\
            \bottomrule
        \end{tabular}%
\end{table}

\begin{table}[t!]
    \centering
    \setlength{\tabcolsep}{3pt}
    \caption{\textbf{The performance of baseline models across scenes of varying difficulty levels.}
    }
    \label{tab:results_level}
    \resizebox{\linewidth}{!}{%
        \begin{tabular}{lrrrrrrrrr}%
            \toprule
            \multirow{2.4}{*}{Baseline} & \multicolumn{3}{c}{Easy} & \multicolumn{3}{c}{Medium} & \multicolumn{3}{c}{Hard} \\
            \cmidrule(lr){2-4} \cmidrule(lr){5-7} \cmidrule(lr){8-10} & SR $\uparrow$ & OSR $\uparrow$ & RDR $\downarrow$ & SR $\uparrow$ & OSR $\uparrow$ & RDR $\downarrow$ & SR $\uparrow$ & OSR $\uparrow$ & RDR $\downarrow$ \\
            \midrule
            ESRP-PPO    & $33.33\%$ & $33.33\%$ & $0.9040$ & $0.54\%$ & $0.54\%$ & $1.0645$ & $0.00\%$ & $0.00\%$ & $1.0048$ \\
            ESRP-BC     & $29.44\%$ & $29.44\%$ & $0.9631$ & $0.91\%$ & $4.20\%$ & $1.1157$ & $0.00\%$ & $4.44\%$ & $1.0704$ \\
            ESRP-VLM     & $27.64\%$ & $27.64\%$ & $0.9523$ & $0.00\%$ & $3.99\%$ & $1.0372$ & $0.00\%$ & $0.00\%$ & $1.0378$ \\
            ESRP-PLAN   & $\textbf{35.73}\%$ & $\textbf{35.73}\%$ & $\textbf{0.7101}$ & $\textbf{22.40}\%$ & $\textbf{39.71}\%$ & $\textbf{0.6425}$ & $0.00\%$ & $\textbf{22.00}\%$ & $\textbf{0.8165}$ \\    
            \bottomrule
        \end{tabular}%
    }%
\end{table}

\noindent\textbf{Varying Difficulty Levels} \label{para: long-horizon}
\Cref{tab:results_level} presents the overall rearrangement performance of our baseline models across scenes of varying difficulty levels. These results reveal that current approaches struggle progressively as the number of objects requiring rearrangement increases. 
This trend reflects compounded challenges, including long-horizon planning under egocentric partial observability, combinatorial object ordering decisions as the number of targets grows, and dynamically evolving scenes that introduce cascading dependencies, all of which make spatial reasoning and planning increasingly difficult as task complexity scales. 
Performance degradation is particularly pronounced in complex scenes, highlighting the \textbf{challenge of maintaining spatial reasoning and planning capabilities as task complexity scales}.

\noindent\textbf{Failure-Mode Decomposition} \label{para: failure_decomposition}
To complement the aggregate metrics with a diagnostic view, we partition each test rollout into four mutually exclusive stages, defined by how far the agent progresses through the rearrangement pipeline (\Cref{tab:stage_analysis}). Across \textit{ESRP-PPO}, \textit{ESRP-BC}, and \textit{ESRP-VLM}, the dominant failure mode is consistently Stage~3 (placement failure), accounting for $60.85\%$, $62.85\%$, and $66.55\%$ of all rollouts, respectively, whereas only $4$--$6\%$ of episodes terminate at the engagement stage and $11$--$13\%$ stall during navigation. This decomposition shows that, regardless of paradigm, the principal bottleneck of \ac{esrp} is \textbf{precise placement under egocentric observability} rather than navigation or grasp initiation, providing a concrete diagnostic target for future research.


\noindent\textbf{Effect of a Stronger VLM Backbone}
To assess whether the modest performance of \textit{ESRP-VLM} is attributable to the specific \ac{vlm} backbone rather than to the ReAct-based prompting paradigm itself, we additionally evaluate \textit{ESRP-VLM} with Gemini~2.5~Flash-Lite~\cite{gemini2.5flashlite} in place of Qwen3-VL-2B-Instruct. Replacing the backbone improves performance across all three metrics, from $17.64\%$ SR, $18.97\%$ OSR, and $0.9830$ RDR with Qwen3-VL-2B-Instruct to $23.01\%$ SR, $23.75\%$ OSR, and $0.9450$ RDR with Gemini~2.5~Flash-Lite. This confirms that stronger general-purpose \ac{vlm} priors partially alleviate the difficulty of \ac{esrp}.

\noindent\textbf{Qualitative Results} \cref{fig:results} and \cref{fig:resultf} illustrate the rearrangement process of \textit{ESRP-PPO}. In \cref{fig:results}, the agent successfully identifies the designated object and moves it to the target location, completing the task within half of the available steps. \cref{fig:resultf} highlights some failure cases of our method. Although the scene is not fully rearranged successfully, the agent achieves a lower \ac{rdr} by placing objects closer to their designated goals.
This suggests that although the agent’s perception is adequate, \textbf{efficiently planning under time constraints and achieving robust performance from egocentric viewpoints remain critical challenges} requiring further research.

\begin{table}[t!]
    \centering
    \setlength{\tabcolsep}{3pt}
    \caption{\textbf{Failure-mode decomposition.}
    Each test rollout is assigned to exactly one of four mutually exclusive stages, defined by how far the agent progresses through the rearrangement pipeline.
    Stage~1: the agent never approaches a rearrangeable object (\textit{navigation failure}).
    Stage~2: the agent reaches an object but never grasps it (\textit{engagement failure}).
    Stage~3: the agent grasps an object but fails to place it at the goal (\textit{placement failure}).
    Stage~4: complete success.
    ESRP-VLM results are based on a single run.}
    \label{tab:stage_analysis}
        \begin{tabular}{lcccc}
            \toprule
            Baseline & Stage 1 (\%) & Stage 2 (\%) & Stage 3 (\%) & Stage 4 (\%) $\uparrow$ \\
            \midrule
            ESRP-PPO & $13.21$          & $5.76$           & $\textbf{60.85}$ & $\textbf{20.18}$ \\
            ESRP-BC  & $\textbf{11.33}$ & $6.73$           & $62.85$          & $19.09$ \\
            ESRP-VLM & $11.64$          & $\textbf{4.18}$  & $66.55$          & $17.64$ \\
            \bottomrule
        \end{tabular}%
\end{table}

\subsection{Discussion}
\label{sec: discussion}

Based on the experimental results, we discuss four key questions regarding the \ac{esrp} task:

\noindent\textbf{Q1: What makes the task challenging, and how can these challenges be addressed?}

Comparing planning- and learning-based approaches in \Cref{tab:results} reveals two main sources of difficulty. The first is \textbf{3D Scene Understanding}: the gap between \textit{ESRP-PLAN} (perfect perception) and learning-based methods (partial egocentric observations) reflects the difficulty of inferring global scene states from local views. The second is \textbf{Long-Horizon Task Planning}: even with perfect perception, \textit{ESRP-PLAN} still fails on Hard-level scenes ($0\%$ SR, \Cref{tab:results_level}), showing that the combinatorial complexity of multi-object rearrangement poses a challenge independent of perception.



\noindent\textbf{Q2: Are the experimental settings and task abstractions reasonable?}

Our design deliberately abstracts low-level control to isolate the core challenges of embodied AI. The high-level manipulation primitives we adopt (``magic snapping''), together with strict collision checks, are shared with mainstream embodied rearrangement benchmarks~\cite{weihs2021visual, szot2021habitat, mirakhor2024task}, and policies trained under such abstractions transfer to physical platforms with only routine sim-to-real adjustments~\cite{yenamandra2023homerobot}. This choice is further validated by \Cref{tab:results_level}: current methods still fail on complex scenes (e.g., $0\%$ \ac{sr} in Hard scenarios) despite the simplified manipulation, confirming that the bottleneck is \textit{long-horizon planning} and \textit{spatial reasoning} under partial observability, not grasping mechanics -- exactly the capability \benchmark is designed to stress-test.

\noindent\textbf{Q3: How does \ac{esrp} relate to real-world deployment, given that all experiments are conducted in simulation?}

Our simulation-only evaluation follows established practice in embodied rearrangement benchmarks~\cite{weihs2021visual, yenamandra2023homerobot}, which favor controlled, reproducible comparisons before real-world deployment. The abstractions \ac{esrp} adopts -- egocentric RGB observations and high-level fetch/release primitives -- are shared with these benchmarks, and prior work shows that policies trained under them transfer to physical platforms with routine sim-to-real adaptation~\cite{yenamandra2023homerobot}. The absence of real-robot results mainly reflects a hardware bottleneck, as commodity platforms cannot yet reliably manipulate furniture-scale objects, whereas the long-horizon planning and partial-observability understanding that \ac{esrp} isolates are exactly the capabilities future deployments will require.

\noindent\textbf{Q4: Is specifying the target layout as a top-down RGB image realistic for real-world deployment?}

Top-down layout images offer an information-dense, unambiguous way to specify multi-object goals: a single image conveys both \textit{which} objects to move and \textit{where} to place them, unlike natural language, which grows lengthy and ambiguous as object count increases. Such images are increasingly easy to acquire: a commodity RGB-D phone or LiDAR tablet can scan a room, reconstruct a 3D mesh, and segment individual furniture pieces; the user then edits the layout (e.g., dragging a sofa) and renders it from a top-down view. This scan--segment--edit--render pipeline is already validated by commercial tools such as MagicPlan~\cite{magicplan}. Beyond user-driven editing, top-down goals can equivalently come from upstream scene synthesis modules~\cite{tang2024diffuscene, luiten2024dynamic}, letting \ac{esrp} operate on either user-authored or generated layouts.

\section{Conclusion}
In this paper, we propose \ac{esrp}, the first embodied rearrangement task that focuses on repositioning furniture within a complex 3D room with the egocentric image as the observation. To accelerate research in this emerging domain, we introduce \benchmark, a comprehensive benchmark that encompasses a large-scale, diverse dataset of indoor scenes and a suite of carefully designed evaluation metrics. By providing both a realistic simulation environment and rigorous performance assessment tools, our work offers a significant step toward advancing embodied rearrangement capabilities.

\noindent\textbf{Limitations}
This work concentrates on room-level furniture rearrangement and relies exclusively on egocentric RGB observations. We envision extending \ac{esrp} to house-level planning and incorporating additional sensor modalities, such as depth, bird's-eye-view cameras, and LiDAR, as vital steps toward broader applicability and real-world deployment.

\bibliographystyle{IEEEtran}
\bibliography{reference}
\end{document}